\documentclass[11pt,a4paper]{article}
\usepackage[hyperref]{acl2018}
\usepackage{times}
\usepackage{latexsym}
\usepackage{soul,color}

\usepackage{tikz}

\usepackage{caption} 
\usepackage{array}
\usepackage{url}

\aclfinalcopy 

\newcolumntype{L}{>{\centering\arraybackslash}m{1.5cm}}

\title{Augmenting word2vec with latent Dirichlet allocation within a clinical application}

\author{Akshay Budhkar \\
  University of Toronto\\
  The Vector Institute \\
  {\tt abudhkar@cs.toronto.edu} \\\And
  Frank Rudzicz \\
  University of Toronto \\
  Toronto Rehabilitation Institute-UHN \\
  The Vector Institute \\
  {\tt frank@cs.toronto.edu}}

\date{}

\begin{document}
\maketitle
\begin{abstract}
This paper presents three hybrid models that directly combine  latent Dirichlet allocation and word embedding for distinguishing between speakers with and without Alzheimer's disease from transcripts of picture descriptions. Two of our models get F-scores over the current state-of-the-art using automatic methods on the DementiaBank dataset.
\end{abstract}

\section{Introduction}
Word embedding projects word tokens into a lower-dimensional latent space that captures semantic, morphological, and syntactic information \citet{mikolov2013distributed}. 
Separately but related, the task of topic modelling also discovers latent semantic structures or \emph{topics} in a corpus. \citet{blei2003latent} introduced latent Dirichlet allocation (LDA), which is based on bag-of-words  statistics to infer topics in an unsupervised manner. LDA considers each document to be a probability distribution over hidden topics, and each topic is a probability distribution over all words in the vocabulary. Both the topic distributions and the word distributions assume distinct Dirichlet priors. 

The inferred probabilities over learned latent topics of a given document (i.e., \emph{topic vectors}) can be used along with a discriminative classifier, as in the work by \citet{luo2014defining}, but other approaches such as TF-IDF \cite{lan2005comprehensive} easily outperform this model, like in the case of the \emph{Reuters-21578} corpus \cite{lewis1987reuters}. To address this, \citet{mcauliffe2008supervised} introduced a supervised topic model, sLDA, with the intention of inferring latent topics that are predictive of the provided label. Similarly, \citet{ramage2009labeled} introduced labeled LDA, another graphical model variant of the LDA, to do text classification. Both these variants have competitive results, but do not address the issue caused by the absence of contextual information embedded in these models.


Here, we hypothesize that creating a hybrid of LDA and word2vec models will produce discriminative features. These complementing models have been previously combined for classification by \citet{liu2015topical}, who introduced topical word embeddings in which topics were inferred on a small local context, rather than over a \emph{complete document}, and input to a skip-gram model. However, these models are limited when working with small context windows and are relatively expensive to calculate when working with long texts as they involve multiple LDA inferences per document.

We introduce three new variants of hybrid LDA-word2vec models, and investigate the effect of dropping the first component after principal component analysis (PCA). These models can be thought of as extending the conglomeration of topical embedding models. We incorporate topical information into our word2vec models by using the final state of the topic-word distribution in the LDA model during training.

\subsection{Motivation and related work}
Alzheimer's disease (AD) is a neurodegenerative disease that affects approximately 5.5 million Americans with annual costs of care up to \$259B in the United States, in 2017, alone \cite{alzheimer20172017}. The existing state-of-the-art methods for detecting AD from speech used extensive feature engineering, some of which involved experienced clinicians. \citet{fraser2016linguistic} investigated multiple linguistic and acoustic characteristics and obtained accuracies up to 81\% with aggressive feature selection. 

Standard methods that discover latent spaces from data, such as word2vec, allow for problem-agnostic frameworks that don't involve extensive feature engineering. \citet{yancheva2016vector} took a step in this direction, clinically, by using vector-space topic models, again in detecting AD, and achieved F-scores up to 74\%. It is generally expensive to get sufficient labeled data for arbitrary pathological conditions. Given the sparse nature of data sets for AD, \citet{noorian2017importance} augmented a clinical data set with normative, unlabeled data, including the Wisconsin Longitudinal Study (WLS), to effectively improve the state of binary classification of people with and without AD.

In our experiments, we train our hybrid models on a normative dataset and apply them for classification on a clinical dataset. While we test and compare these results on detection of AD, this framework can easily be applied to other text classification problems. The goal of this project is to i) effectively augment word2vec with LDA for classification, and ii) to improve the accuracy of dementia detection using automatic methods.

\section{Datasets}
\subsection{Wisconsin Longitudinal Study}
The Wisconsin Longitudinal Study (WLS) is a normative dataset where residents of Wisconsin \emph{(N = 10,317)} born between 1938 and 1940 perform the Cookie Theft picture description task from the Boston Diagnostic Aphasia Examination \cite{goodglass2000boston}. The audio excerpts from the 2011 survey \emph{(N = 1,366)} were converted to text using the Kaldi open source automatic speech recognition (ASR) engine, specifically using a bi-directional long short-term memory network trained to the Fisher data set \cite{cieri2004fisher}. We use this normative dataset to train our topic and word2vec models.

\subsection{DementiaBank}
DementiaBank (DB) is part of the TalkBank project \cite{MacWhinney11}. Each participant was assigned to either the `Dementia' group ($N=167$) or the `Control' group ($N=97$) based on their medical histories and an extensive neuropsychological and physical assessment battery. Additionally, since many subjects repeated their engagement at yearly intervals (up to five years), we use $240$ samples from those in the `Dementia' group, and $233$ from those in the `Control' group. Each speech sample was recorded and manually transcribed at the word level following the CHAT protocol \cite{MacWhinney1992}. We use a $5-$fold group cross-validation (CV) to split this dataset while ensuring that a particular participant does not occur in both the train and test splits. Table \ref{table:db} presents the distribution of Control and Dementia groups in the test split for each fold.

\begin{table}[tbp]
\centering
\resizebox{\columnwidth}{!}{%
\begin{tabular}{|c|c|c|c|c|}
\hline
 & \multicolumn{2}{c|}{\textbf{Sex (M/F)}} & \multicolumn{2}{c|}{\textbf{Age (years)}} \\ \hline
 & AD & CT & AD & CT \\ \hline
WLS & -/- & 681/685 & - (-) & 71.2 (4.4) \\ \hline
DB & 82/158 & 82/151 & 71.8 (8.5) & 65.2 (7.8) \\ \hline
\end{tabular}%
}
\caption{Demographics for DB and WLS for patients with AD and controls (CT). All WLS participants are controls. Years are indicated by their means and standard deviations.}
\label{my-label}
\end{table}

\begin{table}[tbp]
\centering
\resizebox{\columnwidth}{!}{%
\begin{tabular}{|c|c|c|c|c|c|c|}
\hline
\multicolumn{1}{|l|}{} & \multicolumn{1}{l|}{\textbf{Fold 1}} & \multicolumn{1}{l|}{\textbf{Fold 2}} & \multicolumn{1}{l|}{\textbf{Fold 3}} & \multicolumn{1}{l|}{\textbf{Fold 4}} & \multicolumn{1}{l|}{\textbf{Fold 5}} & \multicolumn{1}{l|}{\textbf{Total}} \\ \hline
CT & 55 & 56 & 40 & 40 & 50 & 241 \\ \hline
AD & 56 & 54 & 70 & 70 & 60 & 310 \\ \hline
\end{tabular}%
}
\caption{DB test-data distribution}
\label{table:db}
\end{table}

WLS is used to train our LDA, word2vec and hybrid models that are then used to generate feature vectors on the DB dataset. The feature vectors on the train set are used to train a discriminative classifier (e.g., SVM), that is then used to do the AD/CT binary classification on the feature vectors of the test set.
 
\subsection{Text pre-processing}
During the training of our LDA and word2vec models, we filter out spaCy's list of stop words \cite{spacy2} from our datasets. For our LDA models trained on ASR transcripts, we remove the \emph{[UNK]} and \emph{[NOISE]} tokens generated by Kaldi. We also exclude the tokens \emph{um} and \emph{uh}, as they were the most prevalent words across most of the generated topics. We exclude all punctuation and numbers from our datasets.

\section{Methods}

\subsection{Baselines}
Once an {\bf LDA} model is trained, it can be used to infer the topic distribution on a given document. We set the number of topics empirically to \emph{K=5} and \emph{K=25}. 

We also use a pre-trained {\bf word2vec} model trained on the Google News Dataset \footnote{\texttt{https://code.google.com/archive/p/word2vec/}}. The model contains embeddings for 3 million unique words, though we extract the most frequent 1 million words for faster performance. Words in our corpus that do not exist in this model are replaced with the \emph{UNK} token. We also train our own word vectors with 300 dimensions and window size of 2 to be consistent with the pre-trained variant. Words are required to appear at least twice to have a mapped word2vec embedding. Both models incorporate negative sampling to aid with better representations for frequent words as discussed by \citet{mikolov2013distributed}. Unless mentioned otherwise, the same parameters are used for all of our proposed word2vec-based models.

Given these models, we represent a document by averaging the word embeddings for all the words in that document, i.e.:

\begin{equation}
avg\_word2vec = \frac{\sum\limits_{i=1}^n W_i}{n}
\label{equ:wordvec}
\end{equation}
where $n$ is the number of words in the document and $W_i$ is the word2vec embedding for the $i^{th}$ word. This representation retains the number of dimensions ($N = 300$) in the original model. 

Third, {\bf TF-IDF} is a common numerical statistic in information retrieval that measures the number of times a word occurs in a document, and through the entire corpus. We use a TF-IDF vector representation for each transcript for the top \emph{1,000} words after preprocessing. Only the train set is used to compute the inverse document frequency values. 

Finally, since the goal of this paper is to create a hybrid of LDA and word2vec models, one of the simpler hybrid models -- i.e., {\bf concatenating} LDA probabilities with average word2vec representations -- is the fourth baseline model. Every document is represented by \emph{N + K} dimensions, where $N$ is the word2vec size and $K$ is the number of topics.

\subsection{Proposed models}
\subsubsection{Topic vectors}
Once an LDA model is trained, we procure the word distribution for every topic. We represent a topic vector as the {\em weighted} combination of the word2vec vectors of the words in the vocabulary. This represents every inferred \emph{topic} as a real-valued vector, with the same dimensions as the word embedding. A topic vector for a given topic is defined as:

\begin{equation}
topic\_vector_D = \frac{\sum\limits_{i=1}^{V} p_i W_i}{V}
\label{equ:topvec}
\end{equation}
where $V$ is the vocabulary size of our corpus, $p_i$ is the probability that a given word appears in the topic, from LDA, and $W_i$ is the word2vec embedding of that word.


Furthermore, this approach also represents a given {\em document} (or transcript) using these topic vectors as a linear combination of the topics in that document. This combination can be thought of as a topic-influenced point representation of the document in the word2vec space. A document vector is given by:

\begin{equation}
avg\_topic\_vector_D =  \frac{\sum\limits_{i=1}^{K} p_i T_i}{K}
\end{equation}
where $T_i$ is the topic vector defined in Equation \ref{equ:topvec}, $K$ is the number of topics of the LDA model, and $p_i$ is the inferred probability that a given document contains topic $i$.

\subsubsection{Topical Embedding}
\label{sec:top}

\begin{figure}
\centering
  \includegraphics[width=\linewidth]{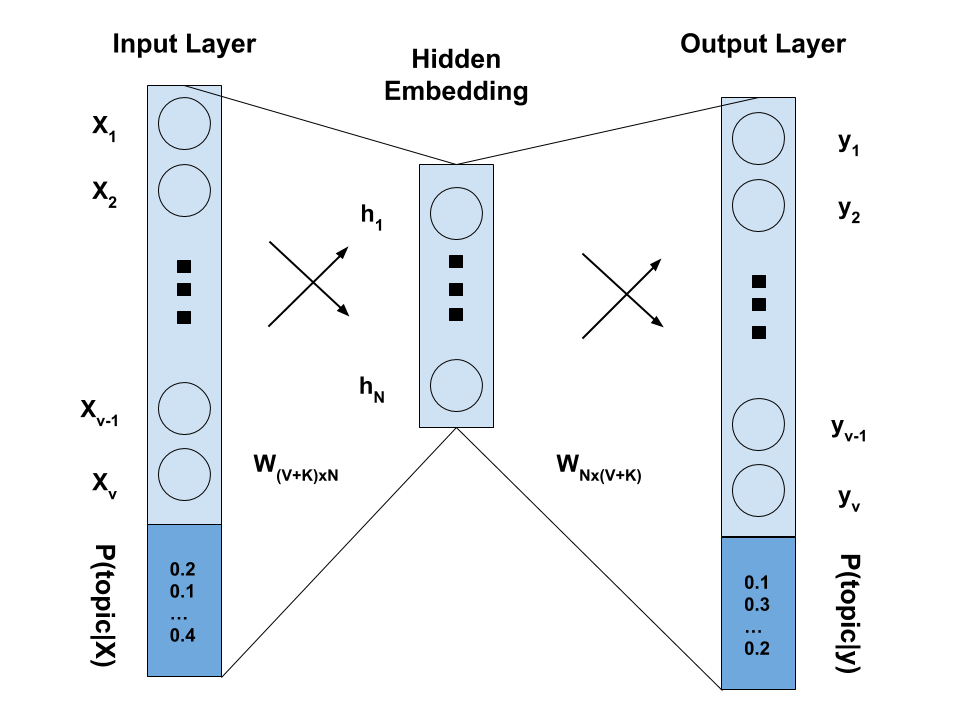}
  \caption{Neural representation of topical word2vec}
  \label{fig:topical}
\end{figure}

To generate topical embeddings, we use the $P(word\,|\,topic)$ from LDA training as the ground truth of how words and topics are related to each other. We normalize that distribution, so that $\sum\limits_{topics} P(topic\,|\,word) = 1$. This gives a topical representation for every word in the vocabulary.

We concatenate this representation to the one-hot encoding of a given word to train a skip-gram word2vec model. Figure~\ref{fig:topical} shows a single pass of the word2vec training with this added information. There, $X$ and $Y$ are the concatenated representations of the input-output words determined by a context window, and $h$ is an $N$-dimensional hidden layer. All the words and the topics are mapped to an $N$-dimensional embedding during inference. Our algorithm also skips the softmax layer at the output of a standard word2vec model, as our vectors are now a combination of one-hot encoding and dense probability matrices. This is akin to what \citet{liu2015topical} did with their LDA inference on a local context document; however, we use the state of the distribution at the last step of the training for all our calculations.

To get document representations, we use the average word2vec approach  in Eq \ref{equ:wordvec} on these modified word2vec embeddings. We also propose a new way of representing documents as seen in Figure \ref{fig:pcaup} where we concatenate the average word2vec with the word2vec representation of the most prevalent topic in the document following LDA inference.

\subsubsection{Topic-induced word2vec}
\label{sec:top2}
Our final model involves inducing topics into the corpus itself. We represent every topic with the string \emph{topic\_i} where $i$ is its topic number; e.g., topic 1 is \emph{topic\_1}, and topic 25 is \emph{topic\_25}. We also create a \emph{sunk} topic character (analogous to \emph{UNK} in vocabulary space) and set it to \emph{topic\_(K+1)}, where $K$ is the number of topics in the LDA model.

\begin{figure}
\centering
  \includegraphics[width=\linewidth]{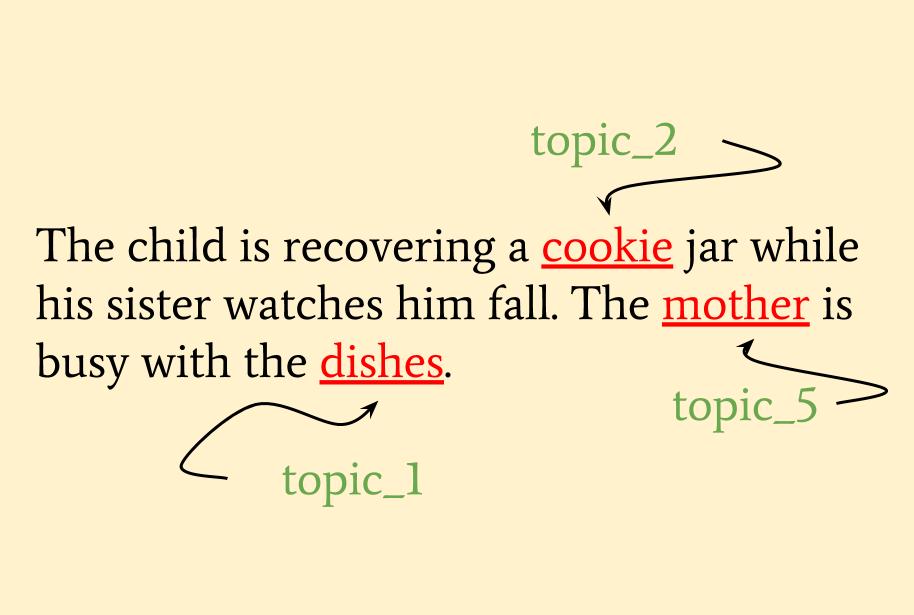}
  \caption{Example topic induction in the WLS corpus.}
  \label{fig:topicinduced}
\end{figure}

We normalize $P(word\,|\,topic)$ to get $P(topic\,|\,word)$ (Section \ref{sec:top}). With a probability of 0.5, set empirically, we replace a given word with the topic string for $\max(P(topic\,|\,word))$, provided the max value is $\geq0.2$. If this max value is $<0.2$, the word is replaced with the sunk topic for that model.

Figure \ref{fig:topicinduced} shows an example of topic induction on a snapshot of an ASR transcript of WLS. This process is repeated $N=10$ times and this augmented corpus is now run through a standard skip-gram word2vec model with dimensions set to 400 to accommodate the bigger corpus. The intuition behind this approach is that it allows words to learn how they occur around $topics$ in a corpus and vice versa.

Document representations follow the same format as in Section \ref{sec:top} where we use the average word2vec vector, and the concatenation of the average word2vec with the word embedding of the most prevalent topic, after LDA inference, as seen in Figure \ref{fig:pcaup}.

\subsection{PCA update}
\label{sec:pca}
\begin{figure}
\centering
  \includegraphics[width=\linewidth]{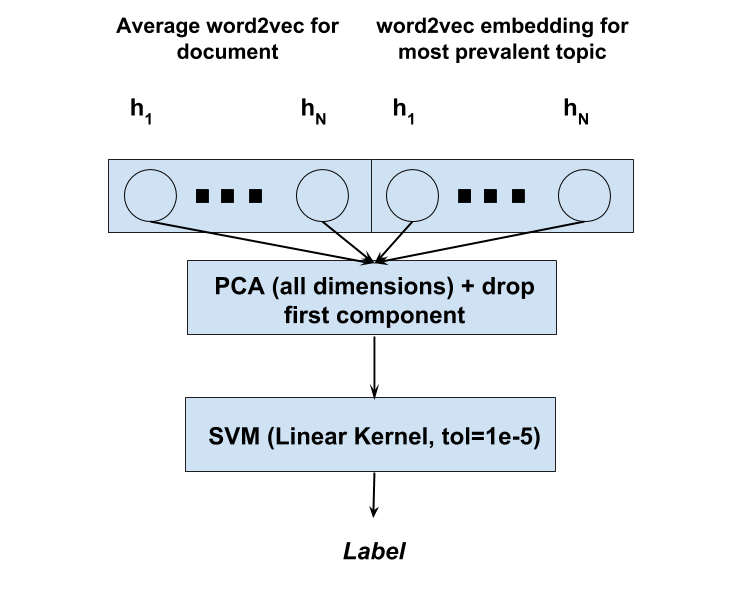}
  \caption{Setup for classification using hybrid models. The PCA step exists for models applying work described in Section \ref{sec:pca}.}
  \label{fig:pcaup}
\end{figure}

Inspired by the work of \citet{arora2016simple}, we transform the features of our models with PCA, drop the first component, and input the result to the classifier. This improves classification, empirically. Figure \ref{fig:pcaup} shows this setup for the Topical Embedding model discussed in Section \ref{sec:top}. 

\subsection{Discriminative classifier}
Apart from the last experiment, where we compare different classifiers on one model, all experiments use an SVM classifier with a linear kernel and tolerance set to $10^{-5}$. All other parameters are set to the defaults in the \texttt{scikit-learn}\footnote{\url{http://scikit-learn.org}} library.

\section{Experimental setup}
\subsection{LDA, word2vec, and hybrid models}
We use Rehurek's Gensim\footnote{\url{https://radimrehurek.com/gensim/}} topic modelling library to generate our LDA and word2vec models. The LDA model follows Hoffman's \cite{hoffman2010online} online learning for LDA, ensuring fast model generation for a large corpus. To train our topical embeddings, we implement the skip-gram variant of word2vec using \texttt{tensorflow}. For all our word2vec models, we set the window size to $2$ and run through the corpus for $100$ iterations.

\subsection{Metric calculation}
We use scikit-learn \cite{pedregosa2011scikit} to classify the vectors generated from our models. For all models, unless specified, we use the default parameters while keeping the discriminative models consistent through all experiments. Our random forest and gradient boosting classifiers each have 100 estimators to be consistent with the work of \citet{noorian2017importance}. We also employ the original pyLDAvis implementation on Github \cite{sievert2014ldavis} to visualize topics across the models. t-SNE \cite{maaten2008visualizing} is used to reduce the vector representations to two dimensions for plotting purposes.

\section{Results}

\subsection{Model Visualization}
\begin{figure}[ht]
\centering
  \includegraphics[width=\linewidth]{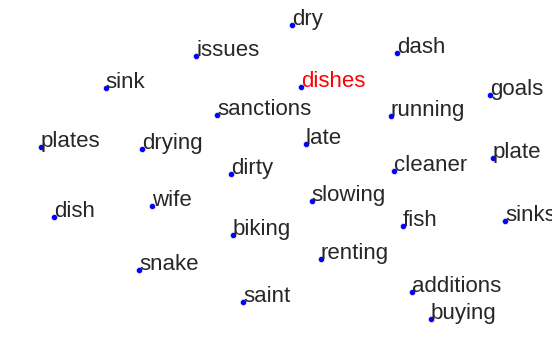}
  \caption{t-SNE reduced visualization for 25 words closest to \emph{dishes} in the Trained word2vec model trained on the WLS corpus.}
  \label{fig:vis1}
\end{figure}

We take the 300-dimensional vector representations of 25 words closest to \emph{dishes} in the word2vec model trained on WLS and run t-SNE dimensionality to plot them on two dimensions in Figure \ref{fig:vis1}. Words similar to \emph{dishes} occur in its vicinity. 

\begin{figure}[ht]
\centering
  \includegraphics[width=\linewidth]{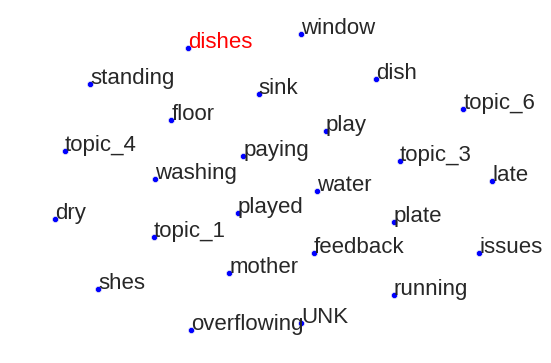}
  \caption{t-SNE reduced visualization for 25 words closest to \emph{dishes} in the Topic-induced LDA-5 model trained on the WLS corpus as discussed in Section \ref{sec:top2}.}
  \label{fig:vis2}
\end{figure}

Figure \ref{fig:vis2} does the same for the topic-induced model (for $K=5$ topics) trained on the augmented corpus as discussed in Section \ref{sec:top2}. In this scenario, we are able to see words \emph{similar} to dishes, and topics that tend to occur in its vicinity. It is evident that all the topics occur close to each other in the embedding space.

\begin{figure*}[ht]
\centering
  \includegraphics[width=\linewidth]{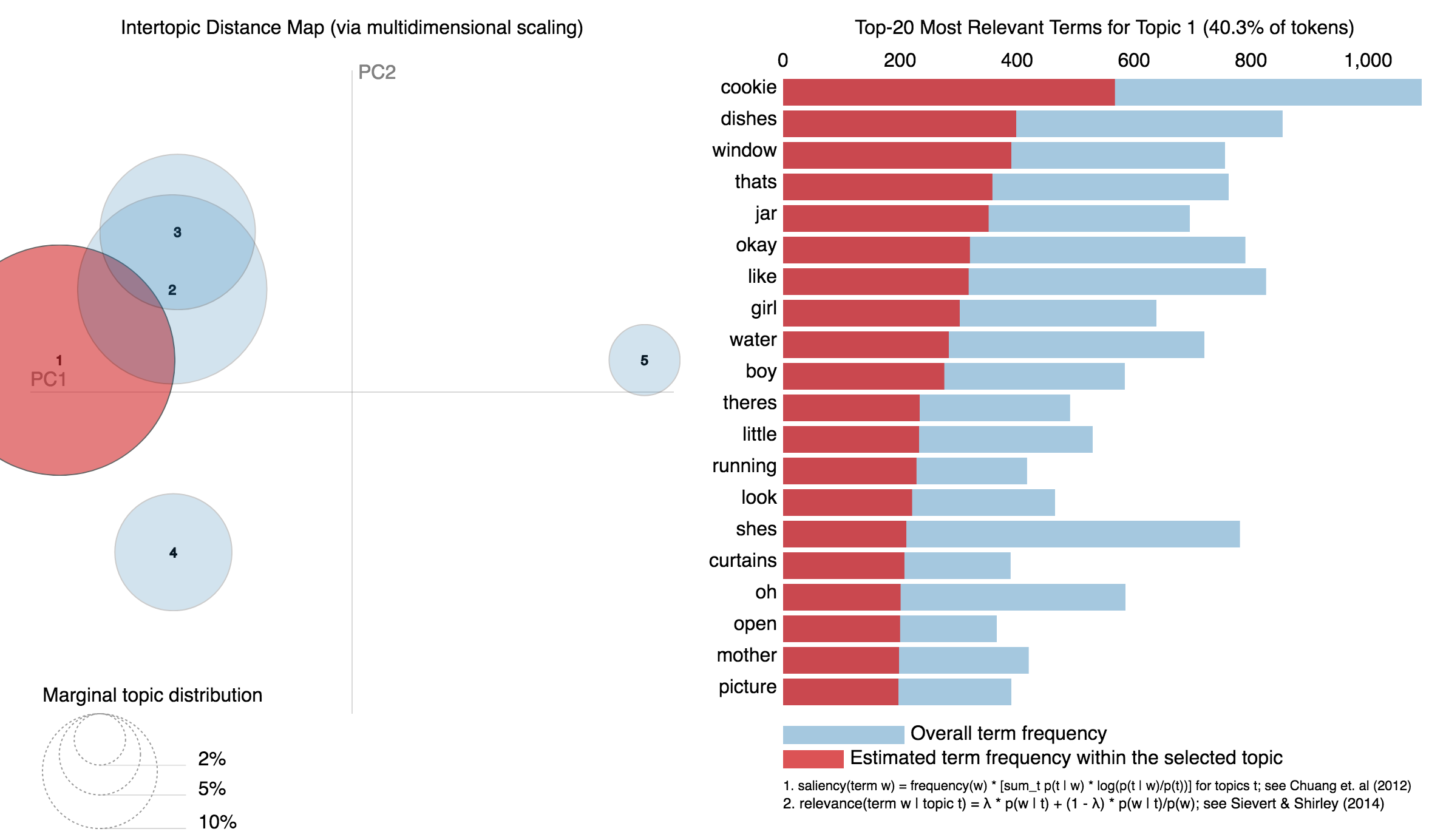}
  \caption{LDA-5 distribution for topic 1. Visual settings of pyLDAvis, and \emph{saliency} and \emph{relevance} were set to default, as provided in \citet{sievert2014ldavis}. PCA was as a dimensionality reduction tool to generate the global topic view on the left. Word distribution on the right is sorted in descending order of the probability of the words occurring in topic 1.}
  \label{fig:pyldavis}
\end{figure*}

Our 5-topic LDA model is visualized using pyLDAvis in Figure \ref{fig:pyldavis}, and the word distribution in topic 1 is shown. Unlike some distinctly varied corpora, like Newsgroup 20 (as seen in \citet{alsumait2009topic}), the topics in WLS do not seem to human-distinguishable and the same few words dominate all the topics. This is expected given that both the AD and CT patients are describing the same picture \cite{goodglass2000boston}, and the top 10 tokens of the stop words-filtered WLS dataset account for 16.89\% of the total words in the corpus.

\subsection{DB classification}
We report the average of the F1 micro and F1 macro scores for the 5-folds for all baseline and proposed models. These results are presented in two parts in Tables \ref{db1} and \ref{db2}.

The LDA-inferred topic probabilities are not discriminative on their own and give an accuracy of $~62.8\%$ (when $K=25$). The TF-IDF model sets a very strong baseline with an accuracy of $74.95\%$, which is already better than the automatic models of \citet{yancheva2016vector} on the same data. Using trained word2vec models for average word2vec representations give better accuracy than using a pre-trained model. Simply concatenating the LDA dense matrix to the average word2vec values gives an accuracy of $74.22\%$ which is comparable to the TF-IDF model. The PCA update on the trained word2vec model boosts the accuracy, and is in line with the work done by \citet{arora2016simple}. This is not the case for the pre-trained word vectors, where the accuracy drops to $59.18\%$ after the update.

The topic vectors end up providing no discriminative information to our classifier. This is the case regardless of whether topic vectors are linearly combined (to get $N=300$ dimensions) or concatenated (to get $N=7500$ dimensions).

\begin{table*}[t]
\centering
\setlength\extrarowheight{5pt}
\resizebox{\textwidth}{!}{%
\begin{tabular}{|c|c|c|c|c|c|c|c|c|c|}
\hline
 & \multicolumn{2}{c|}{\textbf{LDA}} & \multicolumn{2}{c|}{\textbf{Pre-trained word2vec}} & \multicolumn{2}{c|}{\textbf{Trained word2vec}} & \textbf{TF-IDF} & \textbf{Concatenation} & \textbf{Topic Vectors} \\ \hline
 & 5 Topics & 25 Topics &  & PCA Update &  & PCA Update &  &  &  \\ \hline
F1 micro & 55.70\% & 62.78\% & 67.88\% & 59.18\% & 71.50\% & 72.60\% & \textbf{74.95\%} & \textbf{74.22\%} & 56.27\% \\ \hline
F1 macro & 54.44\% & 62.46\% & 66\% & 52.09\% & 71.33\% & 72.25\% & \textbf{74.49\%} & \textbf{73.90\%} & 35.90\% \\ \hline
\end{tabular}%
}
\caption{DB Classification results (Average 5-Fold F-scores): Part 1}
\label{db1}
\end{table*}

\begin{table*}[ht]
\centering
\setlength\extrarowheight{5pt}
\setlength\tabcolsep{5pt}
\resizebox{\textwidth}{!}{%
\begin{tabular}{|c|L|L|L|L|L|L|L|L|L|L|L|L|}
\hline
& \textbf{Topical word2vec} & \textbf{Topical word2vec + topic} & \textbf{Topical word2vec} & \textbf{Topical word2vec + topic} & \textbf{Topic-Induced word2vec} & \textbf{Topic-Induced word2vec + topic} & \textbf{Topic-Induced word2vec} & \textbf{Topic-Induced word2vec + topic} & \textbf{Topic-Induced word2vec} & \textbf{Topic-Induced word2vec + topic} & \textbf{Topic-Induced word2vec} & \textbf{Topic-Induced word2vec + topic} \\ \hline
 & \multicolumn{2}{c|}{25 topics} & \multicolumn{2}{c|}{25 topics and PCA} & \multicolumn{2}{c|}{5 topics} & \multicolumn{2}{c|}{25 topics} & \multicolumn{2}{c|}{5 topics and PCA} & \multicolumn{2}{c|}{25 topics and PCA} \\ \hline
F1 micro & 75.32\% & 75.32\% & 73.69\% & 71.14\% & 75.32\% & 75.68\% & \textbf{77.50\%} & 76.40\% & \textbf{77.10\%} & 74.59\% & \textbf{76.77\%} & 75.31\% \\ \hline
F1 macro & 74.97\% & 75.01\% & 73.32\% & 70.70\% & 74.98\% & 75.36\% & \textbf{77.19\%} & 76.09\% & \textbf{76.86\%} & 72.27\% & \textbf{76.48\%} & 75\% \\ \hline
\end{tabular}%
}
\caption{DB Classification results (Average 5-Fold F-scores): Part 2}
\label{db2}
\end{table*}

The 25-topic topical embedding model discussed in Section \ref{sec:top} outperforms the TF-IDF baseline and gives accuracies of $75.32\%$ when using the average word2vec approach. There is a slight improvement when we concatenate the topic information. All topic-induced models beat the topical embedding model, with the 25-topics variant giving a 5-fold average accuracy of \textbf{$77.5\%$}.

The PCA updates to most of these models decrease the accuracy of classification except for the 5-topic topic-induced variant, where the accuracy increases from $75.32\%$ to $77.1\%$ when using average word2vec as features to a SVM classifier, and from $75.68\%$ to $76.4\%$ when using the concatenated variant.

To check if our accuracies are statistically significant, we calculate our test statistic (\emph{Z}) as follows:

\begin{equation}
\label{eq:z}
Z =  \frac{p_1 - p_2}{\sqrt{2\bar{p}(1-\bar{p})/n}}
\end{equation}

where $(p_1, p_2)$ are the proportions of samples correctly classified by the two classifiers respectively, $n$ is the number of samples (which in our case is $551$) and $\bar{p} = \frac{p_1 + p_2}{2}$. 

Augmenting word2vec models with topic information significantly improves accuracy in the topic-induced word2vec model ($p=0.0227$) when compared to the vanilla-trained word2vec model. This change is not significant, however, in the topical embedding model ($p=0.152$), though it still outperforms \citet{yancheva2016vector}.

\subsection{Ablation study}
Using the best-performing model (i.e., the 25-topic topic-induced word2vec model with average word2vec as features), we consider other discriminative classifiers. As seen in Table \ref{ablation}, the linear SVM model gives the best accuracy of $77.5\%$, though all  other models perform similarly, with accuracies upwards of $70\%$. There is no statistically significant difference between using an SVM vs. a logistic regression ($p=0.569$) or a gradient boosting classifier ($p=0.094$).

\begin{table}[ht]
\centering
\resizebox{\columnwidth}{!}{%
\begin{tabular}{|l|c|c|}
\hline
\multicolumn{1}{|c|}{\textbf{Discriminative Classifier}} & \textbf{F1 micro} & \textbf{F1 macro} \\ \hline
SVM w/ linear kernel & \textbf{77.50\%} & \textbf{77.19\%} \\ \hline
Logistic Regression & 76.05\% & 75.51\% \\ \hline
Random Forest & 71.13\% & 69.97\% \\ \hline
Gradient Boosting Classifier & 73.14\% & 72.39\% \\ \hline
\end{tabular}%
}
\caption{DB: Discriminative Classifiers on Topic-induced LDA-25 model}
\label{ablation}
\end{table}

\section{Discussions}

Although the topic distributions of the LDA models were not distinctive enough in themselves, they capture subtle differences between the AD and CT patients \emph{missed} by the vanilla word2vec models. Simple concatenation of this distribution to the document increases the accuracy by $2.72\%$ ($p=0.31$).

Topic vectors on their own do not provide much generative potential for this clinical data set. The hypothesis is that representing a document as a single point in space, after going through two layers of contraction, removes relevant information to classification.

However, using the same word-topic distribution, normalizing it per word, and combining that information directly into word2vec training increases accuracies $(p=0.152)$. Concatenating the topical embedding to the average word2vec also helps to boost accuracy slightly.

\subsection{Topic-induced negative sampling}
Our novel topic-induced model performs the best among our proposed models, with an accuracy of $77.5\%$ on a 5-fold split of the DB dataset. To put this in perspective, \citet{yancheva2016vector}'s automatic vector-space topic models achieved $74\%$ on the same data set, albeit with a slightly different setup.

The idea of adding topics as strings to the corpus is an idea similar to adding noise during negative sampling of word2vec \cite{mikolov2013distributed}. However, the vanilla word2vec models incorporate negative sampling, and are substantially outperformed by our topic-induced variants. The intuition of letting the words `know' the kind of topics that occur around them, and vice versa, seems to be conducive in incorporating that information into the embeddings themselves. These \emph{noisy-}additions to the corpus also get assigned meaningful embeddings, as can be seen in certain cases where the concatenation model outperforms the average word2vec variant.

Applying PCA to the features does not have a significant trend.

\section{Conclusions}
In this paper, we show the utility of augmenting word2vec with LDA-induced topics. We present three models, two of which outperform vanilla word2vec and LDA models for a clinical binary text classification task. By contrast, topic vector baselines collapse all the relevant information and only perform randomly.

Our topic-induced model with 25 topics trained on WLS and tested on DB achieve an accuracy of $77.5\%$. Going forward, we will test this model on other tasks, diagnostic and otherwise, to see its generalizability. This can provide a starting point for clinical classification problems where labeled data may be scarce.

\section{Acknowledgements}
The Wisconsin Longitudinal Study is sponsored by the National Institute on Aging (grant numbers R01AG009775, R01AG033285, and R01AG041868), and was conducted by the University of Wisconsin.

\bibliographystyle{acl_natbib}
\bibliography{acl2018}

\end{document}